\documentclass[lettersize,journal]{IEEEtran}
\usepackage{array}
\usepackage[caption=false,font=normalsize,labelfont=sf,textfont=sf]{subfig}
\usepackage{textcomp}
\usepackage{stfloats}
\usepackage{verbatim}
\usepackage{cite}
\usepackage{amsmath}
\usepackage{amsfonts}
\usepackage{mathrsfs}
\usepackage{amssymb}
\usepackage{amsthm}
\usepackage{stmaryrd}

\usepackage{algorithm}
\usepackage{algorithmic}
\usepackage{stackengine}
\usepackage{graphicx}
\usepackage{multirow}
\usepackage{epstopdf}
\usepackage{multicol}
\usepackage{caption}
\usepackage{subfig}
\usepackage{enumitem}
\usepackage{amsmath}
\usepackage{pifont}
\usepackage{bbm}
\usepackage{makecell}
\usepackage{mathtools}
\usepackage[table]{xcolor}
\usepackage{url}
\usepackage{bm}
\usepackage{pifont}
\usepackage{stfloats}
\usepackage[T1]{fontenc}
\usepackage{arydshln}
\usepackage{colortbl}
\usepackage{balance}
\usepackage{footmisc}

\usepackage{tikz}

\makeatletter
\newcommand*{\rom}[1]{\expandafter\@slowromancap\romannumeral #1@}
\makeatother

\DeclareMathOperator*{\argmin}{arg\,min}

\usetikzlibrary{positioning,shapes,arrows}

\begin{document}

\title{On the Sparse DAG Structure Learning \\ Based on Adaptive Lasso}

\author{Danru Xu, Erdun Gao, Wei Huang, Menghan Wang, Andy Song, Mingming Gong
\thanks{Danru Xu, Erdun Gao, Wei Huang, and Mingming Gong are with the School of Mathematics and Statistics, Faculty of Science, The University of Melbourne, Melbourne, Australia (e-mail: danru.xu@student.unimelb.edu.au; erdun.gao@student.unimelb.edu.au; wei.huang@unimelb.edu.au; mingming.gong@unimelb.edu.au).
Menghan Wang is with Ebay inc. (email: wangmengh@zju.edu.cn).
Andy Song is with the School of Computing Technologies, the Royal Melbourne Institute of Technology, Melbourne, Australia (e-mail: andy.song@rmit.edu.au).}}

\markboth{Journal of \LaTeX\ Class Files,~Vol.~xx, No.xxx, xxx~xxxx}%
{Shell \MakeLowercase{\textit{et al.}}: A Sample Article Using IEEEtran.cls for IEEE Journals}

\maketitle

\begin{abstract}
Learning the underlying Bayesian Networks (BNs), represented by directed acyclic graphs (DAGs), of the concerned events from purely-observational data is a crucial part of evidential reasoning. This task remains challenging due to the large and discrete search space. A recent flurry of developments followed NOTEARS~\cite{zheng2018dags} recast this combinatorial problem into a continuous optimization problem by leveraging an algebraic equality characterization of acyclicity. However, the continuous optimization methods suffer from obtaining \textit{non-spare graphs} after the numerical optimization, which leads to the inflexibility to rule out the potentially cycle-inducing edges or false discovery edges with small values. To address this issue, in this paper, we develop a completely data-driven DAG structure learning method without a predefined value to post-threshold small values. We name our method NOTEARS with adaptive Lasso~(NOTEARS-AL), which is achieved by applying the adaptive penalty method to ensure the sparsity of the estimated DAG. Moreover, we show that NOTEARS-AL also inherits the oracle properties under some specific conditions. Extensive experiments on both synthetic and a real-world dataset demonstrate that our method consistently outperforms NOTEARS.
\end{abstract}

\section{Introduction}

Bayesian networks (BNs)~\cite{pearl1988probabilistic} is a kind of graphical model that bridges probability theory and graph theory to represent a joint distribution, which can also be leveraged to model the shape and level of the relationships between random variables by representing conditional dependence as edges and corresponding parameters in a directed acyclic graph (DAG). With success in encoding uncertain expert knowledge in expert systems~\cite{heckerman2008tutorial}, BN has been widely used in many applications such as causal inference\cite{spirtes2000causation}, biology~\cite{sachs2005causal}, and healthcare~\cite{lucas2004bayesian}. 

Over the last decades, lots of work has been done on discovering Bayesian networks from purely-observational data. In general, the problem is ill-posed since there may be multiple DAGs producing the same distribution. Furthermore, concerning the computation complexity~\cite{ganian2021complexity}, since the search space of DAGs is discrete, many standard numerical algorithms cannot be utilized, which leads to the DAG structure learning problem NP-hard~\cite{NP_hard, chickering1996learning}. In other words, searching the optimal DAG in the DAGs space scales super-exponentially by the number of variables.

In general, there are mainly two classes of DAG structure learning methods. One is constraint-based algorithms. The other is score-based algorithms. Constraint-based methods, such as SGS~\cite{spirtes1989causality}, PC~\cite{spirtes1991algorithm}, and GS~\cite{margaritis1999bayesian}, read the conditional independence relationships encoded in the data and try to find a completed partially directed acyclic graph (CPDAG) that entails all (and only) the corresponding d-separations by Markovian and Faithfulness assumptions~\cite{peters2017elements}. That is to say, the output of these algorithms is a set of DAGs (equivalence class) instead of the ground-truth DAG. The main idea of score-based methods is to establish a suitable score function as the objective function and then find the graph structure that maximizes the objective in the solution space. Common score functions include BGe score~\cite{geiger1994learning}, BDe score~\cite{heckerman1995learning}, MDL~\cite{bouckaert1993probabilistic} and BIC~\cite{geiger1994learning}. Based on the idea of order search and greedy search, some approaches that optimize the score functions over the DAGs space are proposed, such as order-based search (OBS)~\cite{teyssier2012ordering}, the max-min
hill climbing method~\cite{tsamardinos2006max}, greedy equivalence search (GES)~\cite{chickering2002optimal}, fast GES~\cite{ramsey2017million}, and A*~\cite{ng2021reliable}.

To improve the searching efficiency, recently, Zheng et al.~\cite{zheng2018dags} introduced a new approach, named NOTEARS, which seeks to replace the traditional combinatorial constraint by a smooth characterization of acyclicity whose value is set to be zero as a DAG. In this way, the combinatorial optimization problem is converted into a continuous optimization such that a wide range of numerical methods can be applied. As NOTEARS only considers the linear model, some following works~\cite{kalainathan2018structural, ng2019masked, yu2019dag, lachapelle2020gradient, zhu2020causal, zheng2020learning} extend it to non-linear or non-parametric situations by leveraging neural networks or orthogonal series. Following these works, many works also extend this work to time-series data~\cite{pamfil2020dynotears}, unmeasured common confounders ~\cite{bhattacharya2021differentiable}, interventional data~\cite{brouillard2020differentiable}, distributed learning~\cite{Ng2022federated, gao2022feddag} and incomplete data~\cite{geffner2022deep, gao2022missdag}. Further related works include GOLEM~\cite{ng2020role}, a likelihood-based method that suggests using likelihood as the learning objective with soft constraints. NOFEARS~\cite{wei2020nofears} focuses on optimization by considering the Karush–Kuhn–Tucker (KTT) conditions. Truncated Matrix Power Iteration~\cite{zhang2022truncated} finds that enlarging the coefficients for higher-order terms can better approximate the DAG constraints and leads to a better graph recovery performance. The convergences of the
augmented Lagrangian method and the quadratic penalty
method are studied in~\cite{ng2022convergence}.


\begin{figure}
	\centering
	\includegraphics[width=0.9\columnwidth]{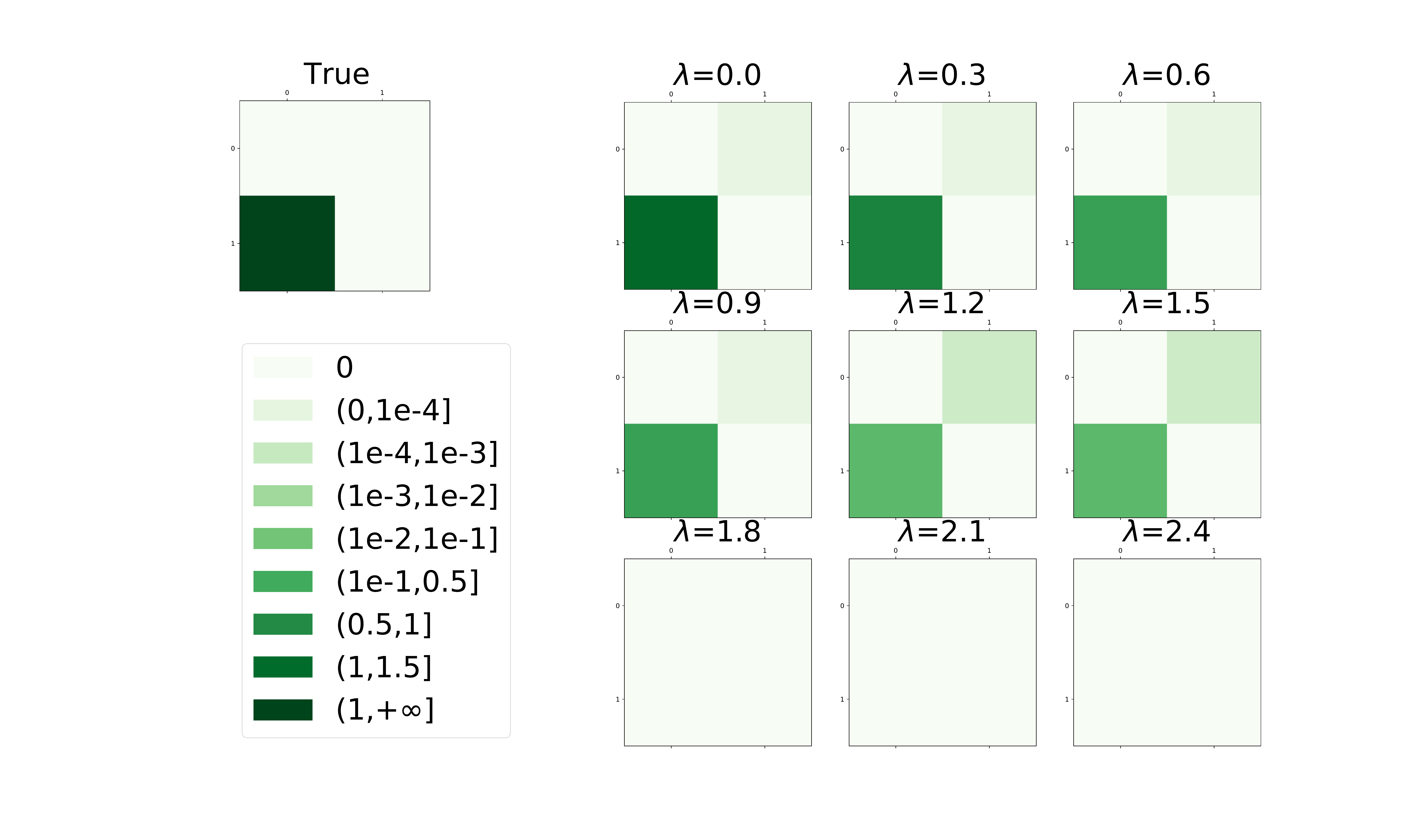} 
	\caption{Visualization of the ground truth (left-hand side of legend) and estimated weighted adjacency matrices (right-hand side of legend) of a $2$-node DAG with $n=1000$ and penalty level $\lambda\in\{0,0.3,0.6,0.9,1.2,1.5,1.8,2.1,2.4\}$. NOTEARS obtains all the solutions without thresholding.}
	\label{fig1}
\end{figure}

It is worth noting that NOTEARS and most of its following works~\cite{zheng2020learning, zhu2020causal, lachapelle2020gradient, yu2019dag, ng2019masked, ng2020role} cannot directly obtain a sparse solution since the numerical optimization always introduces many small false positive estimates. From Figure~\ref{fig1}, we can observe that even with a large sample size over a $2$-node graph, the false positive edge (the right one) always exists and cannot be ruled out by regularization. Moreover, as the penalty level goes larger, the bias of the estimates becomes larger and larger until a pure zeros matrix. This implies that, in NOTEARS, only increasing the value of the penalty coefficient cannot guarantee a consistent estimate, and extra procedures are necessary for obtaining a sparse solution.

To deal with this issue, NOTEARS utilizes a hard threshold to post-process the initial estimators. The fixed thresholding strategy is a direct method to cut down the false positive estimates caused by numerical error. On the other hand, the hard-threshold strategy is not a flexible and systematic way to reduce the number of false discoveries. Firstly, choosing a fixed and sub-optimal value as a threshold cannot adapt to different types of graphs. Secondly, although the amount of false positive edges is diminished via adopting a hard-thresholding strategy, small coefficients in the true model may also be deleted indiscriminately (see Section~\ref{subsec:neighbour0}).

For NOTEARS, on the one hand, it cannot obtain a sparse solution without thresholding. On the other hand, threshold strategy causes inflexibility, and this is not a systematic way. Motivated by this, we develop a new DAG structure learning method based on the adaptive lasso, which abandons the fixed-threshold strategy and directly leads to a sparse output. As we will show, the main contributions are as follows.

\begin{enumerate}
	\item{} We propose a purely data-driven method that does not need a predefined threshold and then illustrate the specific algorithm.
	\item{} We study the oracle properties that our method possesses, including sparsity and asymptotic normality, which means that our method has the ability to select the right subset model and converges to the true model with an optimal rate when the sample size increases.
	\item{} We consider using a more sufficient and suitable validation set in cross-validation for our goal to find the optimal subset model.
	\item{} We demonstrate the effectiveness of our method through simulation experiments with weights generated from a broader range, e.g., Gaussian distributed weights with mean zero.
\end{enumerate}

The rest of this paper is organized as follows: in Section~\ref{sec:background}, we review some basic concepts of the DAG model and the specific algorithm of NOTEARS, as well as the limitation of its fixed-threshold strategy. Then, based on the idea of adaptive Lasso~\cite{zou2006adaptive}, we introduce our method and study its asymptotic behavior when the sample size goes to infinity in Section~\ref{sec:method}. In Section~\ref{sec:experiments}, to verify our theoretical study and the effectiveness of our method, we compare our method with NOTEARS and existing state-of-the-art by simulated experiments. Finally, we conclude our work in Section~\ref{sec:conclusion}.
\section{Related works}
\label{sec:background}

In this section, we review the linear DAG model and the NOTEARS method. In section~\ref{subsec:linearDAG}, we first briefly review some fundamental concepts of the DAG model with linear structure assumption and the corresponding score function used to maximize the likelihood of observations. Then, we detail the NOTEARS method and its optimization algorithm in Section~\ref{subsec:fix_thres}.

\subsection{DAG with Linear Structure}
\label{subsec:linearDAG}

\begin{figure}[h]
	\centering
	\includegraphics[width=\columnwidth]{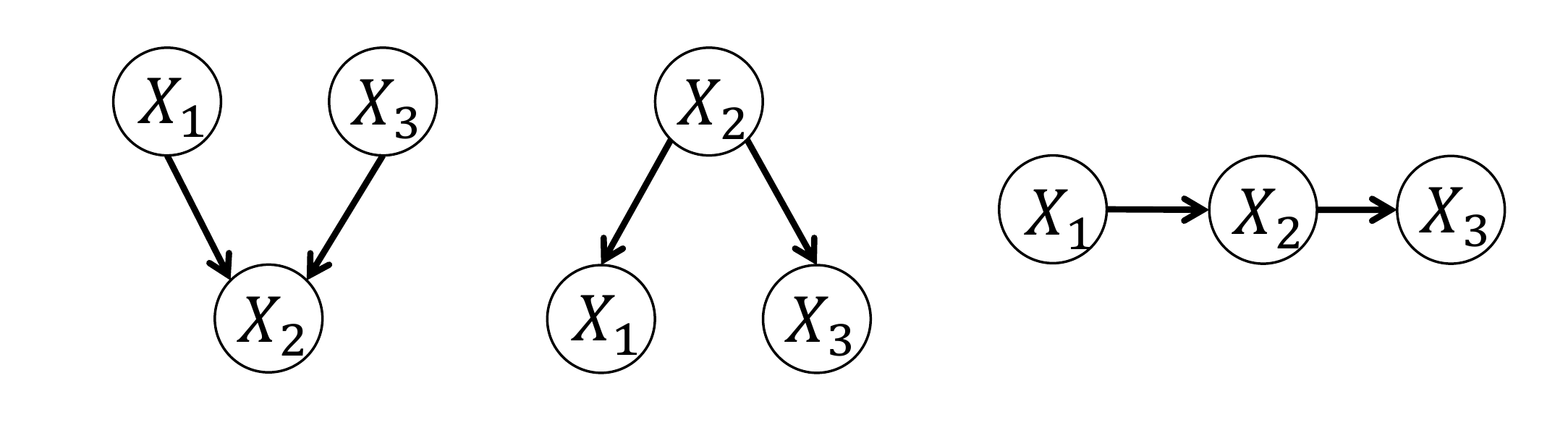} 
	\caption{No circle in these three graphs. Both $X_1$ and $X_3$ are parental nodes of $X_2$ (left); $X_2$ is parental node of $X_1$ and $X_3$ (middle); $X_2$ is the parent of $X_3$ and the child of $X_1$.}
	\label{fig2}
\end{figure}

A DAG can be encoded as $G=(V, E)$, where $V$ refers to the set of nodes and $E$ is the set of directed edges. The nodes in $G$ represent variables under consideration. And directed edges correspond to the direct conditional relationships between parental nodes and child nodes. For example, $V_1\rightarrow V_2$, represents $V_1$ is the parent of $V_2$. Figure~\ref{fig2} illustrates three possible three-node DAGs with two edges.

The task of DAG structure learning is to seek a DAG from the DAGs space of $d$ nodes (discrete) that best fits the observational data $\mathbb{X}$. Let $\mathbb{X}\in \mathbb{R}^{n \times d}$ represent the observations consisting of $n$ $d$-dimensional vectors. We assume that these vectors are realizations of $n$ identically independently distributed (iid) random vectors with joint distribution $P(X)$, where $X=(X_1, X_2,..., X_d)$. Importantly, with no latent variable and Markovian assumptions~\cite{spirtes2001causation}, the joint distribution $P(X)$ can be decomposed as follows:
\begin{equation}
    P(X)=\prod_{i=1}^{d}P(X_i|Pa(X_i)),
\end{equation}
where $Pa(X_i)$ represents the parental set of $X_i$. Then, we model the relationships of variables in $X$ by a structure equation model~(SEM), denoted as $X_i=f_i(Pa(X_i),\varepsilon_i), i=1,...,d$, where $\varepsilon_i$ refers to the uncovered variables. In this paper, we narrow our focus on the DAG with additive noise models (ANMs). By introducing the concept of weighted adjacency matrix $W\in \mathbb{R}^{d\times d}$, we can establish the linear ANMs as $X = XW + N$, where $N$ is a $d$-dimensional random vector, denoted as $N=(N_1,N_2,...,N_d)$. Each $N_i$ represents the noise term added to $X_i$. In this paper, we assume $N_1, N_2,..., N_d$ are independent and identically distributed. The information about the structure of a DAG is completely involved in $W$, which is defined in the following way:
\begin{equation}
    \{V_i \rightarrow V_j\} \subset E \quad \Longleftrightarrow \quad W_{ij} \ne 0.
\end{equation}

We note $G_W$ as the graph induced by $W$. To derive $W$, we can maximize the likelihood of observations $\mathbb{X}$. Benefiting from the well-known probability density transformation method~\cite{ng2020role}, this target can be equally achieved by minimizing the least-square loss function $\frac{1}{2n}||X-XW||_2^2$ over the search space. A regularization term $\lambda||W||_1$ is added to deduce a sparse solution. Then, the overall score function is 
\begin{equation}
	\label{equ1}
	\mathcal{S}(W)=\frac{1}{2n}||X-XW||_2^2+\lambda||W||_1.
\end{equation}

\subsection{DAG with no NOTEARS}
\label{subsec:fix_thres}

Previous score-based methods suffer from searching the discrete DAGs space and lead to low efficiency. To overcome this problem, Zheng et al.\cite{zheng2018dags} proposed the NOTEARS method, which seeks to use a function $h$ satisfying these four desiderata: i) $h(W)=0$ is a necessary and sufficient condition for $G_W$ to be a DAG; ii) its value reflects the level of acyclicity of the graph; iii) smooth; iv) its first derivative exists and is easy to compute. Based on the method in~\cite{Harary1971}, which starts to use the matrix algebra to investigate the number of circles in a graph, NOTEARS verifies that $h(W) = \text{Tr}(e^{W\circ W}) - d $ is a proper function to characterize the acyclicity with the derivative $\nabla h(W)=(e^{W\circ W})^T\circ2W$. After determining the smooth equality constraint, the problem can be written as the following optimization program:
\begin{equation}
	\label{equ2}
	\begin{aligned}
	\min_{W\in \mathbb{R}^{d\times d}} & \ \frac{1}{2n}||X-XW||_2^2+\lambda||W||_1, \\
	\text{subject to} & \ h(W)=0.
	\end{aligned}
\end{equation}

In order to solve the program (\ref{equ2}), this constrained optimization program is converted to a sequence of unconstrained problems by the augmented Lagrangian method \cite{aug_Lagrangian}. The procedure is shown as follows.
\begin{equation}
	\label{equ3}
	\begin{aligned}
		W_k= & \ \argmin_{W\in \mathbb{R}^{d\times d}} \frac{1}{2n}||X-XW||_2^2+\lambda||W||_1 \\
		&\quad \quad \quad \quad +\frac{\rho_k}{2}|h(W)|^2 +\alpha_k h(W),\\
		\alpha_{k+1}= & \ \alpha_k + \rho_k h(W_k), \\
		\rho_{k+1} = & \ \beta \rho_k.
	\end{aligned}
\end{equation}

For each iteration, the first step can be solved by the L-BFGS proximal quasi-Newton method~\cite{byrd1995limited, Proximal_QN} with the derivative as
$-\frac{1}{n}X^T(X-XW)+\lambda \mathbbm{1}_{d\times d}+\left(\rho|h(W)|+\alpha_k\right)\nabla h(W).$ To overcome the problem caused by the non-smooth point zero in the penalty term, i.e., $W_{ij}=0,\quad i,j=1,..,d$, the original weighted adjacency matrix is cast into the difference between two matrices $(W^1, W^2)$ with non-negative elements \cite{box_constraint}. 

\textbf{Remark} After obtaining the initial estimate, NOTEARS zeros out small coefficients if their absolute values are less than the predefined threshold to reduce the number of false positive discovery edges. However, this strategy leads to more missing edges with small weights and is hard to adapt to different scenarios. For example, if the weights of a DAG are generated from a uniform distribution over $(0,2)$, by setting $w_t=0.3$, the estimates of small weights, which are located in the interval $(0,0.3)$, have a high risk to be forced to $0$ in the post-processing step. Since the simulation experiments in previous work commonly set a gap to zero, for instance, in the simulation part of \cite{zheng2018dags}, the weights of edges commonly sampled uniformly from $[-2\alpha,-0.5\alpha]\cup [0.5\alpha, 2\alpha]$, where $\alpha$ is a weight scale, this side effect is not obvious. The specific algorithm of NOTEARS with a fixed predefined threshold is presented in Algorithm~\ref{alg:alg1}.

\begin{algorithm}[htbp]
	\caption{NOTEARS with a fixed threshold.}\label{alg:alg1}
	\begin{algorithmic}
		\STATE 
		\STATE \textbf{Step 0.} Take $\rho_0 > 0$, initial value $W_0\in \mathbb{R}^{d\times d}$, $\alpha_0\in \mathbb{R}$, and some $\xi\in(0,1)$. Set threshold $w_{t}=0.3$, $\eta=10$.
		\STATE \textbf{Step k.} (k$\ge$1) Find the smallest non-negative integer $j_k$ such that with $\rho=\eta^{j_k} \rho_0$
		\[
		h(W_{k+1})<\xi h(W_k),
		\]
		where
		\begin{equation*}
			\begin{aligned}
				W_{k+1}=\argmin_{W\in \mathbb{R}^{d\times d}}&\frac{1}{2n}||X-XW||_2^2+\lambda||W||_1\\
				&+\frac{\rho}{2}|h(W)|^2+\alpha_k  h(W).
			\end{aligned}
		\end{equation*}
		Compute $\alpha_{k+1}=\alpha_{k}+\rho h(W_{k+1})$ \\
		$\rho_{k+1} = \beta \rho_k$
		\STATE 
		\STATE Set $\widehat{W} = W_{k+1}$ if converge.
		\STATE \textbf{Final Post-processing step.} Set $\widehat{W}\left[|\widehat{W}|<w_{t}\right]=0$, then return $\widehat{W}$.
	\end{algorithmic}
	\label{alg1}
\end{algorithm}

\section{The proposed method}
\label{sec:method}
	
In this section, we extend the NOTEARS method with adaptive Lasso to achieve learning the sparse DAG structure from purely observational data and name our method as NOTEARS with adaptive Lasso (NOTEARS-AL). We first introduce the statistical tool we use in our method, named adaptive lasso, in Section~\ref{subsec:adap_lasso}. After detailing our method in Section~\ref{subsec:method}, we investigate the asymptotic behavior of NOTEARS-AL theoretically in Section~\ref{subsec:Asymptotic}. We then develop an algorithm to implement this method in Section~\ref{subsec:algorithm}, and illustrate how we deal with our model's hyper-parameters in Section~\ref{subsec:haper-para}. Finally, We introduce how to extend NOTEARS to generalized linear models in Section~\ref{subsec:extension}.

\subsection{Adaptive Lasso}
\label{subsec:adap_lasso}
In recent decades, variable selection has been studied extensively. In particular, Lasso \cite{tibshirani1996regression} was introduced to improve the prediction accuracy and, simultaneously, reduce overfitting. In the linear model, the lasso estimates are obtained by optimizing the following program:
\begin{equation}
\min_{\beta \in \mathbb{R}^{d}} \ \frac{1}{2n}||X-X\beta||_2^2+\lambda_n||\beta||_1.
\end{equation}
where $\lambda_n$ is related to the sample size $n$. Although Lasso variable selection has been shown to be consistent under some conditions \cite{meinshausen2006variable}, Zou\cite{adaptive_lasso} illustrated certain scenarios where the lasso is an inconsistent procedure in terms of variable selection. Zou\cite{adaptive_lasso} argued that the setup in lasso is somewhat unfair since it treats all the coefficients with indistinctive penalty levels in the sparsity term.
To fix this problem, Zou\cite{adaptive_lasso} proposed a new methodology called adaptive lasso. In adaptive lasso, the weights of penalty levels are totally data-driven and cleverly chosen. The adaptive lasso estimates are obtained by optimizing the following program:
\begin{equation}
\min_{\beta \in \mathbb{R}^{d}} \frac{1}{2n}||X-X\beta||_2^2+\lambda_n\sum_{i=1}^{d}c_{i}|\beta_i|,
\end{equation}
where $c_i$ is a predefined penalty weight. For example, we can use the ordinary least square estimates $\widehat{\beta_{ols}}$ to define $c_i$. Choose a $\gamma > 0$, and define the penalty weight $c_i=\frac{1}{|\widehat{\beta_{ols, i}}|^{\gamma}}$, where $\gamma$ commonly be $0.5$, $1$, $2$ or determined by the cross-validation method.

\subsection{The Overall Learning Objective}
\label{subsec:method}

From Figure~\ref{fig1}, we can find that when the penalty level is relatively low, the learned false edges cannot be zeroed out. On the other hand, assigning a larger weight to the regularization term may lead to even worse results. This implies that \textit{Lasso put much more penalty on large coefficients rather than false positive ones, which is not consistent with our requirement}. Motivated by this observation, to improve the criterion (\ref{equ2}) on variable selection, one intuition is to apply the adaptive penalty levels to different coefficients, which has been developed in~\cite{zou2006adaptive}. Following this motivation, we modify Eq.~(\ref{equ2}) to
\begin{equation}
	\label{equ5}
	\begin{aligned}
		 \min_{W\in \mathbb{R}^{d\times d}}  & \ \frac{1}{2n}||X-XW||_2^2+\lambda_n\sum_{i=1}^{d}\sum_{j=1}^{d}|c_{ij}w_{ij}|, \\
		 \text{subject to} &  \ h(W)=0,
	\end{aligned}	
\end{equation}
where $c_{ij}$ is the specified penalty for $W_{ij}$. Expecting the minor false positive edges can be shrunk to zeros and reserve the true edges at the same time, we would like the corresponding $c_{ij}$ to be large for the former and small for the latter. Hence, the minor false edges are heavily penalized, and the true positive edges are lightly penalized. More details about how to choose $c_{ij}$ are given in Section~\ref{subsec:Asymptotic}. With this modification of the score function, no extra predefined threshold is needed to post-processing the initial estimate anymore.

\subsection{Asymptotic Oracle Properties}
\label{subsec:Asymptotic}

In this section, we show that with a proper choice of $c_{ij}$ and under some mild conditions, our method possesses the oracle properties when the sample number $n$ goes to infinity. Similar to procedure (\ref{equ3}), via the augmented Lagrangian method, a dual ascent process can solve the ECP (\ref{equ5}). The first step of each iteration can be regarded as finding the local minimum of the Lagrangian with a fixed Lagrange multiplier $\alpha$, that is,
\begin{equation}
	\label{equ6}
	\small
	\begin{aligned}
	L_n(W)=\min_{W\in \mathbb{R}^{d\times d}}&\frac{1}{2n}||X-XW||_2^2+\lambda_n\sum_{i,j}|c_{ij}w_{ij}|\\
		&+\frac{\rho}{2}|h(W)|^2+\alpha h(W).
	\end{aligned}
\end{equation}

Let $W^*$ denote the underlying true adjacency matrix. And then, we define
\begin{equation*}
\begin{aligned}
    \mathcal{A}=\{(i,j):w^*_{ij}\ne 0\}, \\ \mathcal{A}_c=\{(i,j):w^*_{ij}= 0\},
\end{aligned}
\end{equation*}
which means that $\mathcal{A}$ collects the indices for terms whose true parameters are nonzero, and $\mathcal{A}_c$ contains the indices for terms that do not exist in the underlying true model. Here, we first make two conditions:
\begin{enumerate}
	\item{} $X_i=X_iW^*+N_i, i=1,2,...,n$, where $N_1,N_2,...,N_n$ are identically independent distributed with mean $\bold{0}$ and variance $\mathbb{\sigma}^2<\infty$.
	
	\item{} $\frac{1}{n}X^TX$ is finite and converges to a positive definite matrix $D\in \mathbb{R}^{d\times d} $. 
\end{enumerate}

Under these two conditions, the oracle properties of NOTEARS-AL are described in the following theorems. Rigorous proofs are given in Appendix~\ref{append_proof}.

\noindent \textbf{Theorem 1} (Local Minimizer)  Under above two conditions, let $a_n=\lambda_n c_{\mathcal{A}}$, where $c_{\mathcal{A}}$ refer to the $c_{ij}$'s associated with nonzero coefficients in $W^*$, if $a_n=o_p(n^q)$ for some $q\le-\frac{1}{2}$, then there exists a local minimizer of $L_n(W)$, which is $\sqrt{n}$-consistent to $W^*$. Let $\widehat{W}_n$ denotes the local minimizer in Theorem 1, which satisfies $||\widehat{W}_n-W^*||=O_p(n^{-\frac{1}{2}})$. Consistency is a nice start, and this motivates us to study more on its asymptotic properties as sample size $n$ goes to infinity. Before introducing the oracle properties, we need to define more notations because now we want to analyze the estimators of nonzero parameters in the underlying true model. We define
\begin{equation*}
    \begin{aligned}
         \mathcal{A}_i &=\{(i,j):(i,j)\in \mathcal{A}\}, \\ 
         d_i &=|\mathcal{A}_i|.
    \end{aligned}
\end{equation*}

In condition 1), we already assume $\frac{1}{n}X^TX \to D$ as $n\to \infty$. Furthermore, without loss of generality, we assume $\mathcal{A}_i=\{(i,1),(i,2),...,(i,d_i)\}$. Let
\[D=
\left(\begin{matrix} 
	D_{i0} &D_{i1} \\
	D_{i2}  & D_{i3}  
\end{matrix}\right),
\]
where $D_{i0}$ is a $d_i\times d_i$ matrix corresponding to the coefficients with index in $\mathcal{A}_i$.

\noindent \textbf{Theorem 2} (Oracle Properties) Under the two conditions, let $b_n=\lambda_n c_{\mathcal{A}_c}$, where $c_{\mathcal{A}_c}$ refer to the $c_{ij}$'s associated with zero coefficients in $W^*$, if $\sqrt{n}b_n\rightarrow \infty$ as $n\rightarrow \infty$, then the local minimizer $\widehat{W}_n$ in Theorem 1 must satisfy the following:	
\begin{itemize}
	\item (Sparsity) $ \lim\limits_{n\to \infty}P(\widehat{W}_{\mathcal{A}_c}=0)=1$
	\item (Asymptotic normality) For $\forall i\in{1,2,...,d}$, $\sqrt{n}(\widehat{W}_{\mathcal{A}_i}-{W^*}_{\mathcal{A}_i})\xrightarrow{d}X\sim N(0,\sigma^2 D_{i0}^{-1})$ as $n\rightarrow \infty$.
\end{itemize}

Firstly, for the zero coefficients part, Theorem 2 shows that given $a_n=o_p(n^q)$ for some $q\le-\frac{1}{2}$ and $\sqrt{n}b_n\rightarrow \infty$, NOTEARS-AL can consistently remove all irrelevant variables with probability tending to 1. Secondly, by magnifying the difference by $\sqrt{n}$ for the nonzero estimators, we can see that the pattern turns out to be a normal distribution. Then, there is only one question left. That is how to determine $c_{ij}$'s such that the nonzero related part $a_n=o_p(n^q)$ for some $q\le-\frac{1}{2}$ and zero related part $\sqrt{n}b_n\rightarrow \infty$. Here, using the fact that ordinary least square (OLS) estimates $\widehat{W}_{ols}$ are $\sqrt{n}$-consistent estimates of $W$ (the process of proof is consistent with Theorem1), suppose that $\lambda_n \sqrt{n}\rightarrow 0$ and $\lambda_n n^{\frac{\gamma+1}{2}}\rightarrow \infty$, then, we can define $c_{ij}$'s by $\widehat{W}_{ols}$ with some specific power $\gamma$ so that the above two requirements are met. Besides, specific details to determine $\lambda_n$ are given in \ref{subsec:haper-para}.
\begin{equation}
	\label{equ7}
	\small
	\begin{aligned}
		\widehat{W}_{ols} = & \ \argmin_{W\in \mathbb{R}^{d\times d}}\frac{1}{2n}||X-XW||_2^2+ \lambda_n\sum_{i=1}^{d}\sum_{j=1}^{d}|c_{ij}w_{ij}|, \\
		    \text{subject to} & \ h(W)=0, \ c_{ij} =\frac{1}{|\widehat{W}_{ols_{ij}}|^{\gamma}}.
	\end{aligned}
\end{equation}

\subsection{NOTEARS with Adaptive Lasso}
\label{subsec:algorithm}

The algorithm can be divided into two parts. The first part is to find the solution $\widehat{W}_{ols}$ for Eq.~(\ref{equ7}), and the second part is plugging in the adaptive penalty parameters to Eq.~(\ref{equ6}) and solving it. These equality-constrained programs (ECPs) are then solved by the augmented Lagrangian method, i.e., using the Lagrangian method to handle the hard DAG constraint $h(W)=0$. The first part follows the same procedure as the original NOTEARS without the thresholding step. And then, we obtain a matrix $C\in \mathbb{R}^{d\times d}$ with element $c_{ij}=\frac{1}{|\widehat{W}_{ols_{ij}}|}$. Let $W_C \vcentcolon = C \circ W$, where $\circ$ is the Hadamard product. Thus, we have $W_{ij}=W_{C,ij}/c_{ij}$. In matrix notation, it is $W = W_C \varoslash C$, where $\varoslash$ refers to Hadamard division. Then, ECP (4) can be transformed as 
\begin{equation}
	\label{equ8}
	\small
	\begin{aligned}
		\widehat{W_C}=&\argmin_{W_C\in \mathbb{R}^{d\times d}}\frac{1}{2n}||X-XW_C \varoslash C||_2^2+\lambda_n||W_{C}||_1 \\
		&\text{subject to} \quad h(W_C \varoslash C)=0.\\
		\widehat{W}_{als} = &\widehat{W_C} \varoslash C,
	\end{aligned}
\end{equation}
where $\widehat{W}_{als}$ is the final estimate for $W$. Then, similar to (\ref{equ3}), via the augmented Lagrangian method, the above ECP~(\ref{equ8}) can be solved by a dual ascent process.
\begin{equation}
	\label{equ9}
	\small
	\begin{aligned}
		W_{Ck}=&\argmin_{W_C\in \mathbb{R}^{d\times d}}\frac{1}{2n}||X-XW_C \varoslash C||_2^2+\lambda_n||W_{C}||_1\\
		&+\frac{\rho}{2}|h(W_C \varoslash C)|^2+\alpha_k h(W_C \varoslash C),\\
		\alpha_{k+1}=&\alpha_k+\rho h(W_C \varoslash C).
	\end{aligned}
\end{equation}

To complete the above optimization, we need to derive a new form of the gradient for the first step in Eq.~(\ref{equ9}). Here we also reformulate the original $W$ as the difference between two matrices with positive elements. 
For the Loss function:
\begin{equation*}
	\nabla\frac{1}{2n}||X-XW_C \varoslash C||_2^2=-2X^T(X-XW_C\varoslash C)\varoslash C.
\end{equation*}
For the acyclicity term $h$:
\begin{equation*}
	\nabla h(W_C \varoslash C)=2\left(e^{(W_C\circ W_C)\varoslash (C\circ C)}\right)^T W_C \varoslash  (C\circ C).
\end{equation*}
To sum up, the algorithm proceeds as Algorithm~\ref{alg:alg2}.
\begin{algorithm}[htbp]
	\caption{NOTEARS with the adaptive lasso.}\label{alg:alg2}
	\begin{algorithmic}
		\STATE 
		\STATE \textbf{Step 0.} Take $\rho_0 > 0$, initial value $W_0\in \mathbb{R}^{d\times d}$, $\alpha_0\in \mathbb{R}$, and some $\xi\in(0,1)$. Set threshold $w_{t}$, $\eta=10$.
		\STATE \textbf{OLS loop k.} (k$\ge$1) Using the complete data set $X$, find the smallest non-negative integer $j_k$ such that with $\rho=\eta^{j_k} \rho_0$
		\[
		h(W_{k+1})<\xi h(W_k),
		\]
		where
		\[W_{k+1}=\argmin_{W\in \mathbb{R}^{d\times d}}\frac{1}{2n}||X-W^TX||_2^2+\frac{\rho}{2}|h(W)|^2+\alpha_k  h(W).
		\]
		Compute $\alpha_{k+1}=\alpha_{k}+\rho h(W_{k+1})$.
		\STATE 
		\STATE Set $\widehat{W}_{ols} = W_{k+1}$ if converge. Then, define $C \vcentcolon=1\varoslash|\widehat{W}_{ols}|^{\gamma}$, $W_C \vcentcolon = C\circ W$.
		\STATE
		\STATE \textbf{Adaptive lasso loop t.} (t$\ge$1) Using the complete data set $X$, for all $\lambda_n$, find the smallest non-negative integer $i_t$ such that with $\rho=\eta^{i_t} \rho_0$
		\[
		h(W_{C,t+1}\varoslash C)<\xi h(W_{C,t}\varoslash C),
		\]
		where
		\begin{equation*}
			\begin{aligned}
				W_{C,t+1}=&\argmin_{W_C\in \mathbb{R}^{d\times d}}\frac{1}{2n}||X-XW_C \varoslash C||_2^2+\lambda_n||W_{C}||_1\\
				&+\frac{\rho}{2}|h(W_C \varoslash C)|^2+\alpha_t h(W_C \varoslash C).
			\end{aligned}
		\end{equation*}
		Compute $\alpha_{t+1}=\alpha_{t}+\rho h(W_{C,t+1}\varoslash C)$.
		\STATE Set $\widehat{W_C}_{,n} = W_{C,t+1}$ if converge.
		\STATE Return the adaptive estimate $\widehat{W}_n \vcentcolon=\widehat{W_C}_{,n}\varoslash C$.
	\end{algorithmic}
\end{algorithm}

\subsection{Hyper-parameter}
\label{subsec:haper-para}

Choosing the tuning parameter $\lambda_n$ is an important issue. We found that the results of NOTEARS-based methods are sensitive to the value of hyper-parameter, thus it is essential to choose a good one. Our goal is to choose the optimal model $s_n$ which contains the true variables as many as possible. Here, following Zou~\cite{adaptive_lasso}, we utilize cross-validation to find an optimal $\lambda_n$. From NOTEARS-AL, we can obtain a set of candidate models $S=\{s_n,n=1,2,..., N\}$, where $s_n=\{(i,j)\in (1:d)\times(1:d):\widehat{W}_{n,ij}\ne0\}$. The cross-validation procedure we use shows as Algorithm~\ref{alg:alg3}.
\begin{algorithm}[htbp]
	\caption{The Cross-validation method.}\label{alg:alg3}
	\begin{algorithmic}
		\STATE 
		\STATE \textbf{Step 0.} Divide the data into validation set $X_v$ and train set $X_t$, and $|X_v|=n_v$, $|X_t|=n_t$, $n_v+n_t=n$.
		\STATE \textbf{For model $S_n$ (n=1,2,...,N).}
		Using the validation set $X_t$, compute the solution $\widetilde{W}_n^t$, where
		\[
		\widetilde{W}_n^t=\argmin_{W\in \mathbb{R}^{d\times d},W_{(-s_n)}=0}\frac{1}{2n_t}||X_t-W^TX_t||_2^2,
		\]
		
		\STATE 
		\STATE Evaluate the prediction performance of $\widetilde{W}_n^t$ on the validation set $X_v$ by loss function $\frac{1}{2n_v}||X_v-(\widetilde{W}_n^t)^TX_v||_2^2$.
		
	\end{algorithmic}
	
\end{algorithm}

Another key modification is that we need a more sufficient validation set to find the best model to achieve the restricted model-selection consistency, especially when the candidate model set is large\cite{CV_n}. Motivated by this, instead of using the traditional $K$-fold cross-validation method, whose validation set is only $1/K$ of data, we suggest swapping the proportions of the validation set and training set.

\subsection{Extension to Generalized Linear Models}
\label{subsec:extension}

Zheng et al.\cite{zheng2020learning} also extend the idea of NOTEARS to generalized linear models (GLMs). Traditionally, a GLM is formulated as follows:

\begin{equation}
	\label{glm}
	\mathbb{E}[X_i|Pa(X_i)]=g_i(d_i^TX),
\end{equation}
where $g_i$ is a link function used in GLMs, and $d_i \in \mathbb{R}^{d}$. To extend the DAG constraint beyond the linear setting, the key point is to find a sensible way to redefine the weighted adjacency matrix $W$ since there is no explicit form of $W$ in the GLMs setting. Based on the idea in \cite{rosasco2013nonparametric}, which encodes the importance of each variable considered via utilizing corresponding partial derivatives, it is then easy to show that the dependence among variables can be precisely measured by the $L^2$-norm of corresponding partial derivatives~\cite{zheng2020learning}. This implies the weighted adjacency matrix $W$ in GLMs can be defined as follows:

\begin{equation}
	\label{glm_W}
	W_{ki}=\vert\vert\frac{\partial f_i}{\partial x_k}\vert\vert_{L^2}=d_{ik}^2.
\end{equation}

Thus, for linear mean functions, i.e. $f_i(X) = d_i^TX$, $W_{ki}=0$ is equivalent to $d_{ik}=0$. We can simplify the DAG constraint $h(W)=0$ by plugging in $W=({d_1}^T,{d_2}^T,...,{d_d}^T)$. To construct the score function for GLMs, we need a more suitable loss function here. In this paper, we focus on one specific model in GLMs, the logistic regression model, where $X_i\in\{0,1\},i=0,...,d$. For logistic regression, the least-square loss is not appropriate anymore since it will result in a non-convex optimal function. Here we introduce the concept named Log Loss, which is commonly used as the loss function for logistic regression, defined as follows:
\begin{equation}
\label{equ10}
l =\frac{1}{n}\sum_{i=1}^{n}[-x_i \log\left(g(t_i) -(1-x_i)\log(1-g(t_i)\right)],
\end{equation}
where $t_i=x_i\times ({d_1}^T,{d_2}^T,...,{d_d}^T)$.
\section{Experiments}
\label{sec:experiments}

\noindent We conduct experiments with a large sample size to verify our asymptotic oracle properties (Section~\ref{subsec:Asymptotic}). To validate the effectiveness of our proposed method, we compare it against NOTEARS with fixed threshold, greedy equivalent search (GES) \cite{chickering2002optimal, ramsey2017million} and the PC algorithm \cite{spirtes2000causation}. For NOTEARS, the value of the threshold is $0.1$, and set the penalty level $\lambda=0.1$ in our experiments. This setting can not only promise a DAG result but also remain a relatively good performance. 

The basic setup of our experiments is as follows. For each experiment, the ground-truth DAG is generated from the random graph model: Erd\"os-R\'enyi (ER). We use ER$2$ to denote an ER graph with $s_0 = 2d$ edges and similar meanings for ER$1$ and ER$4$. Based on the ground truth DAG, we generate the weights of existing edges from $1$) a normal distribution with mean $0$ and variance $\sigma^2$; 2) a uniform distribution over the interval $(-c,c)$, where $c\in \mathbb{R}$. And then, we simulate the sample $X$ via $X_i = f_i (X_{pa(i)}) + z_i,\quad i=1,...,d$, and the noise $z_i \sim N(0, 1)$. Here, we mainly consider the linear case and extend to one type of GLMs named the logistic regression model. The metrics we use to evaluate the estimated DAGs include structural Hamming distance (SHD), true positive rate (TPR), and false discovery rate (FDR).

\subsection{Numerical Results: Gaussian Weights}
Compared to the original weight-generating interval of NOTEARS, which is Uniform $([-2.0,-0.5]\cup[0.5,2.0])$, there are two points we consider to improve in our method. The first one is about the upper and lower bound, which are set to be $\pm 2\alpha$. We expect our method can extend NOTEARS to a wider range with a better estimate. The second one is the coefficients near $0$, which cannot be learned by the fixed-thresholding post-processing step. For the above two improvement points, we first perform the structure learning study in Gaussian-weight cases, where the weights of existing edges are sampled from a normal distribution with mean $0$ and variance $2^2$. In this way, we can investigate the performance of NOTEARS-AL under a wider range of coefficients.

\begin{figure}[!t]
	\centering
	\includegraphics[width=0.85\columnwidth]{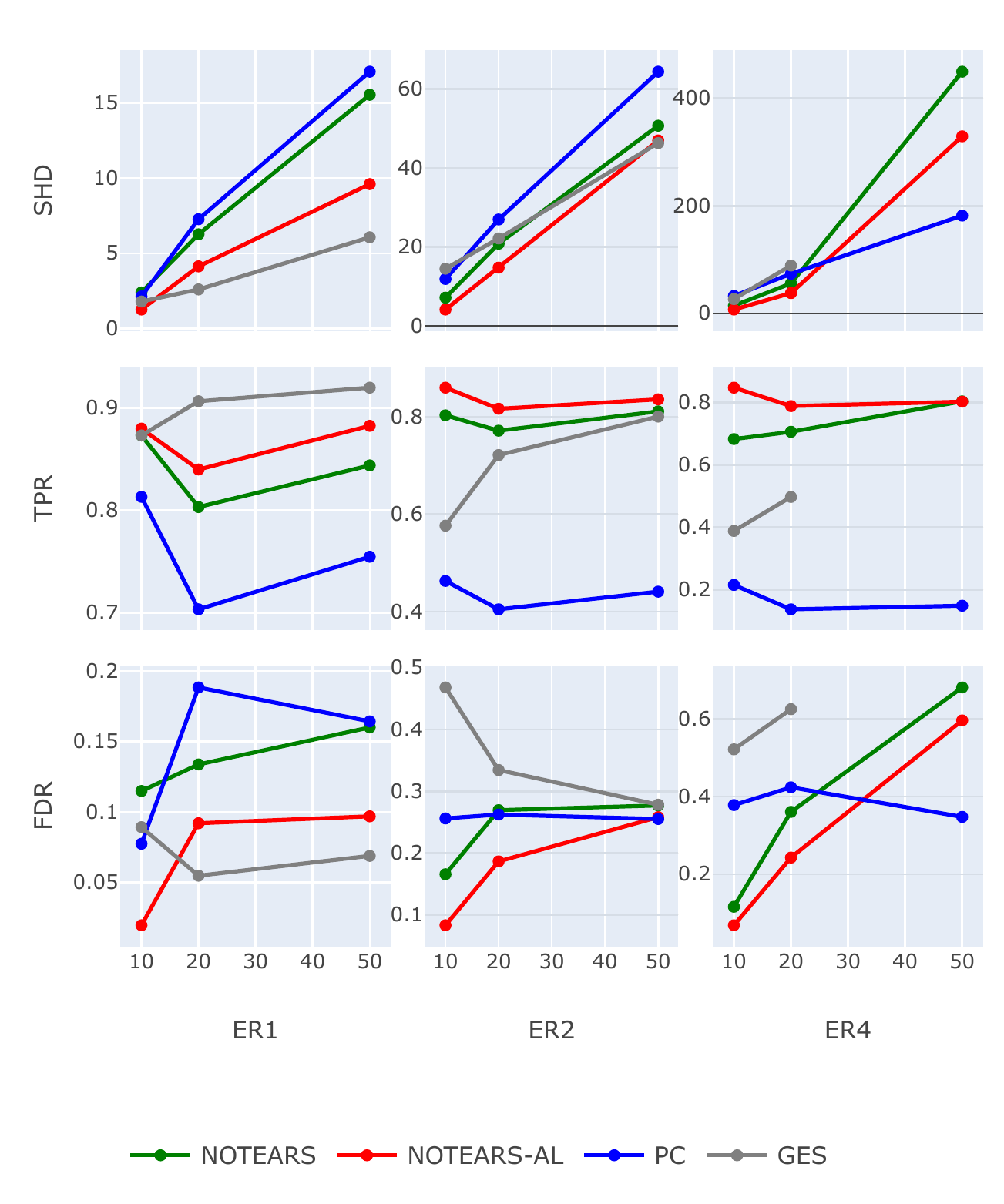} 
	\caption{Numerical results in terms of SHD($\downarrow$), TPR($\uparrow$), and FDR($\downarrow$) on ER graphs, with sample size $n = 1000$ and Gaussian noise. For TPR, higher is better. For the other criterion, lower is better. Columns: random graph types, ER$k$ means ER graphs with $k\times d$ expected edges.} 
	\label{fig3}
\end{figure}

In our experiments, we generate \{ER$1$, ER$2$, ER$4$\} graphs with different numbers of nodes $d=\{10,20,50\}$, $15$ graphs for each type. And then, we simulate $n=1000$ samples with iid Gaussian noises. The results are shown in Figure~\ref{fig3}. For better visualization, only the average values of metrics are reported. Consistent with previous work GES, it shows a significant advantage in the ER$1$ case, no matter which metric we evaluate. But the performances of GES deteriorate rapidly when the level of sparsity decreases. Also, as we can see from Figure~\ref{fig3}, for ER$4$ with $50$ nodes, the algorithm of GES takes too long to obtain the results because of the large searching space and low searching efficiency. Not surprisingly, the performance of NOTEARS and NOTEARS-AL show a positive relationship. This is natural since NOTEARS-AL extracts the weights of penalty level from conventional NOTEARS, which means that the performance of NOTEARS-AL is also sensitive to the results of the conventional one. Nonetheless, our approach performs uniformly better than NOTEARS across different graphs in terms of all metrics.
    
\begin{figure}[!t]
	\centering
	\includegraphics[width=0.85\columnwidth]{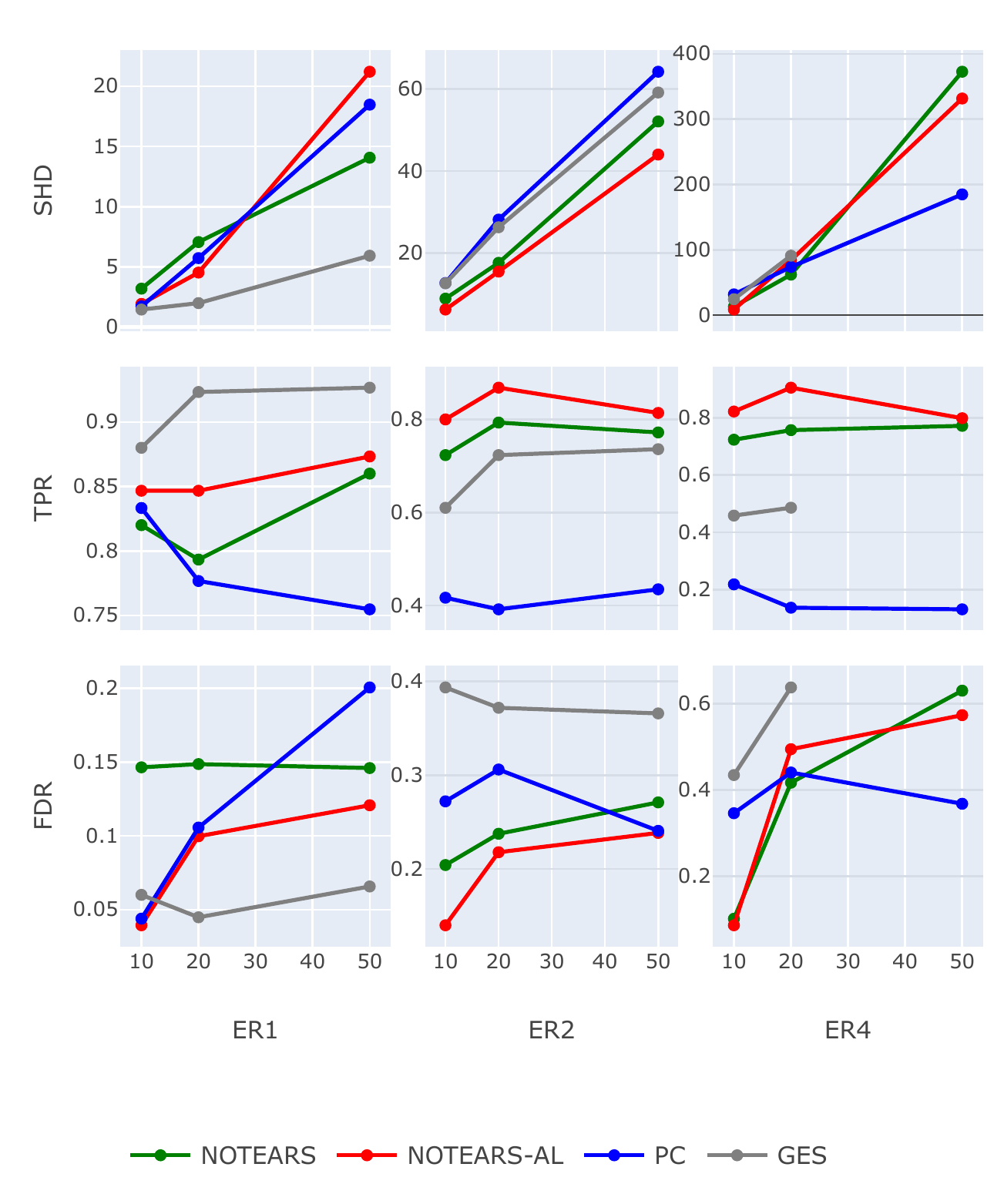} 
	\caption{Numerical results in terms of SHD, TPR, and FDR on ER graphs, with sample size $n = 1000$ and Exponential noise.}
	\label{fig7}
\end{figure}

\begin{figure}[!t]
	\centering
	\includegraphics[width=0.85\columnwidth]{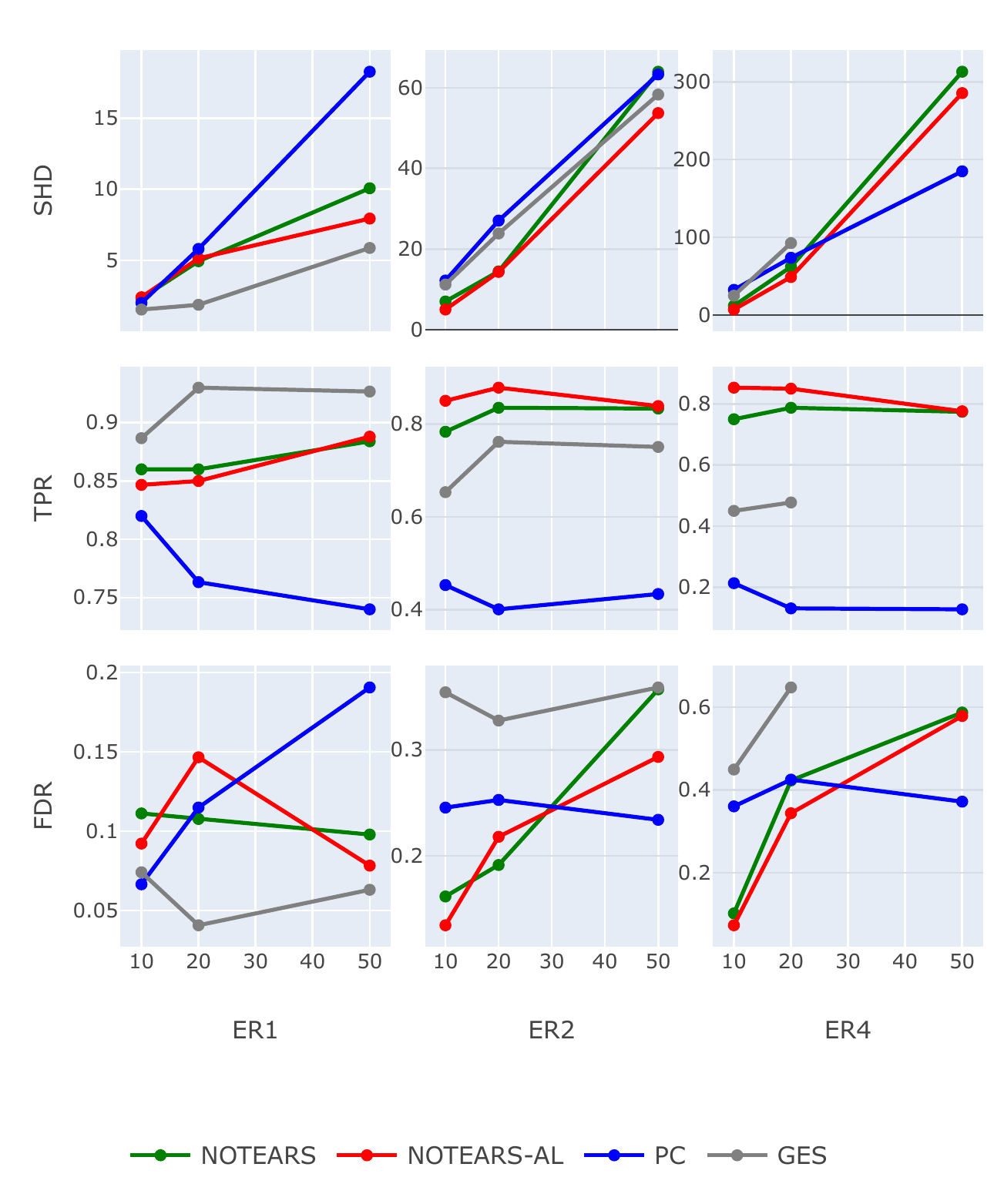} 
	\caption{Numerical results regarding SHD, TPR, and FDR on ER graphs, with sample size $n = 1000$ and Gumbel noise.}
	\label{fig8}
\end{figure}

Figure~\ref{fig7} and Figure~\ref{fig8} show structure recovery results for SEM with exponential noise and Gumbel noise, respectively, where the weights of existing edges are sampled from a normal distribution with mean $0$ and variance $2^2$. 

\subsection{Coefficients in the Neighborhood of $0$}
\label{subsec:neighbour0}
\begin{figure}[htbp]
	\centering
	\includegraphics[width=0.85\columnwidth]{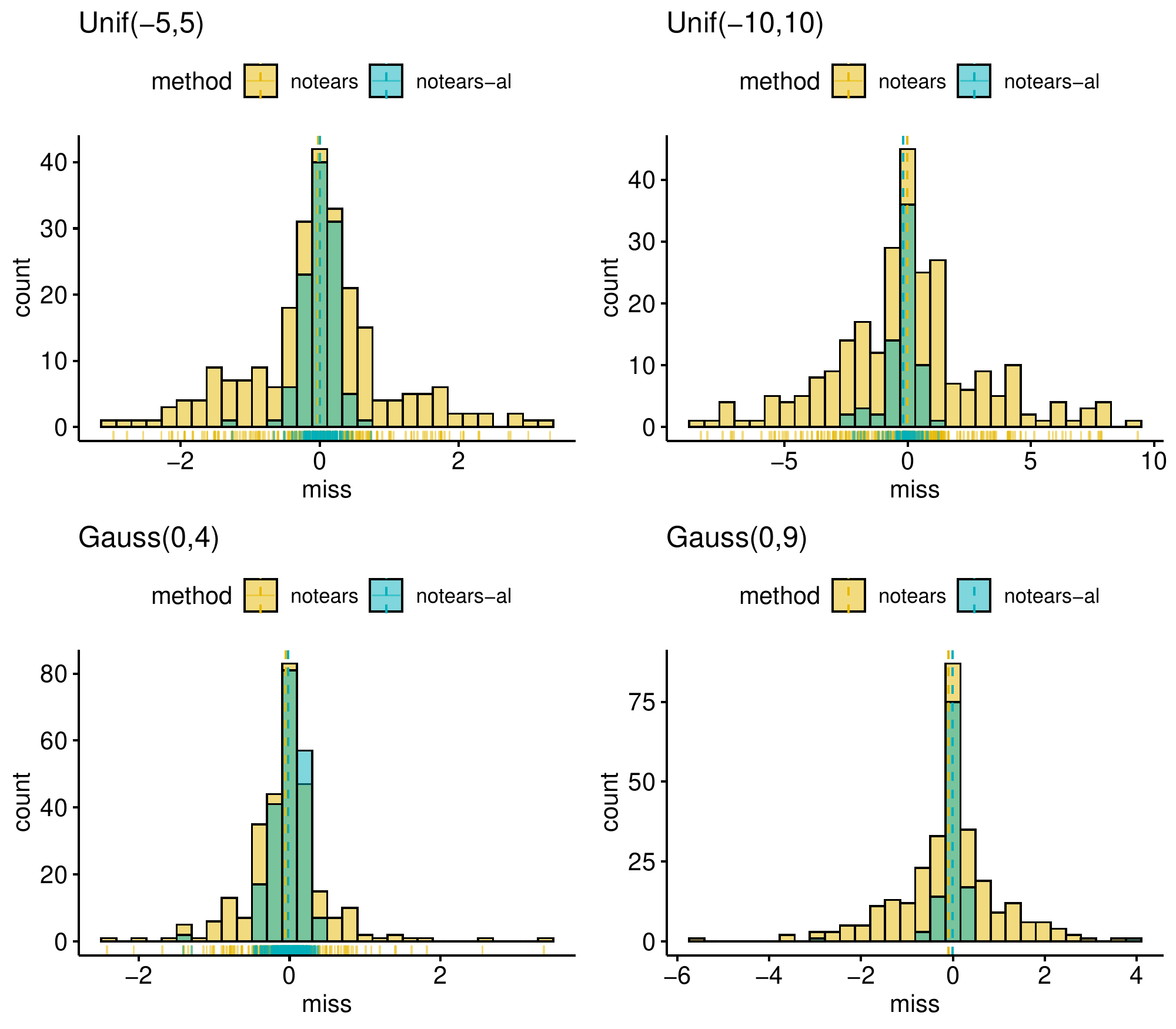} 
	\caption{The histogram of missing edges. Blue bars refer to NOTEARS with a hard-threshold $0.1$; yellow bars refer to NOTEARS-AL. In the first row, the coefficients of the ground truth graph are generated from the uniform distributions $(-5,5)$ and $(-10,10)$ from left to right. In the second row, the true coefficients are from normal distributions with variances $2^2$ and $3^2$, respectively.}
	\label{fig4}
\end{figure}

As we mentioned, one side effect of the fixed-threshold strategy is that small coefficients in the true model may be deleted indiscriminately. Since the simulation experiments in previous work commonly set a gap to zero, which is $\alpha[-2,-0.5]\cup \alpha[0.5,2]$, this side effect is not obvious. Hence, we investigate the value of missing edges when weights are generated from a range that covers zero, e.g., Uniform $(-5,5)$, Uniform $(-10,10)$, Gaussian $(0,2^2)$ and Gaussian $(0,3^2)$. For each kind of distribution, we generated $200$ DAG structures from the ER$1$ model with $10$ nodes and then simulated $1000$ samples. Based on the different methodologies of these two methods, one with a hard threshold of $0.1$ and the other with adaptive penalty levels, our method is more flexible and suitable to handle the coefficients around zero. Indeed, Figure \ref{fig4} confirms this intuition. We plot the histogram of missing edges. In all three cases, our method (blue bar) loses fewer edges than NOTEARS with hard thresholding (yellow bar), especially the area around zero. Moreover, as the range becomes wider or the variance increases, the gap between these two methods becomes more and more significant.

\subsection{High-dimension Cases}
\begin{figure}[htbp]
	\centering
	\includegraphics[width=1.0\columnwidth]{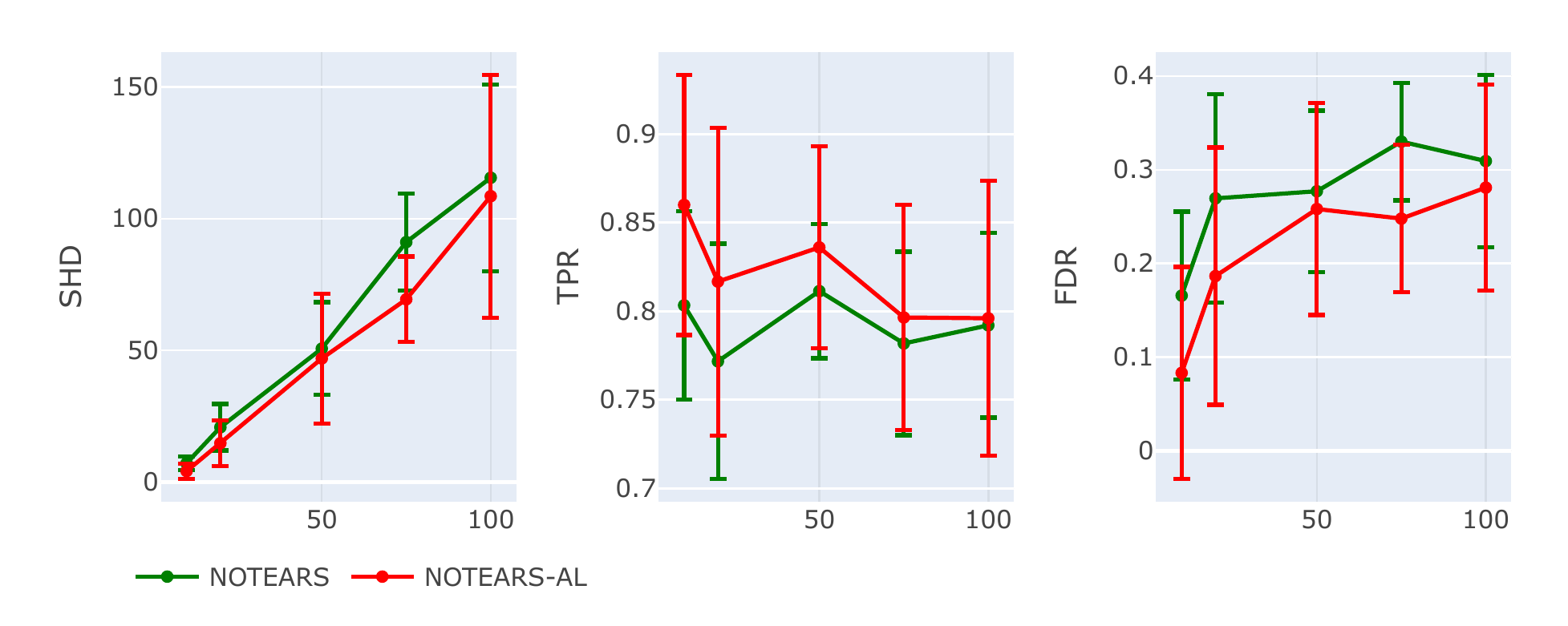} 
	\caption{Numerical results in terms of SHD, TPR, and FDR on ER$2$ graphs with Gaussian $(0,2^2)$ weights, with sample size $n=1000$ and iid Gaussian noise.}
	\label{fig5}
\end{figure}

To investigate the performance of our method for estimating DAG from high-dimensional graphical models, we increase the number of nodes to $100$ and compare the results to those of fixed-thresholding NOTEARS. Here, to illustrate the validity of the evaluation criteria, we draw error bars with sample standard deviations. The results are shown in Figure \ref{fig5}. We can find that under nearly the same value of SHD, the true positive rate of NOTEARS-AL is significantly higher than that of NOTEARS with a fixed threshold. Besides, in terms of FDR (lower is better), NOTEARS-AL has lower values for all graphs. In general, NOTEARS-AL consistently outperforms the original fixed-threshold one, even in high-dimensional settings. 

\subsection{Generalized Linear Models}
For GLMs, we concentrate on logistic regression for dichotomous data $X_i\in\{0,1\}$, $i=1,...,d$ with link function $g(t)=e^t/(e^t+1)$. As we introduced in Section~\ref{subsec:extension}, for linear mean functions, the elements of $W$ in DAG constraint $h(W)=0$ can be substituted by $d_{ik}$'s, the coefficients in linear mean functions. We compare the performance of fixed-thresholding NOTEARS to NOTEARS-AL with edge weights following Gaussian $(0,2^2)$.  

\begin{figure}[htbp]
	\centering
	\includegraphics[width=1.0\columnwidth]{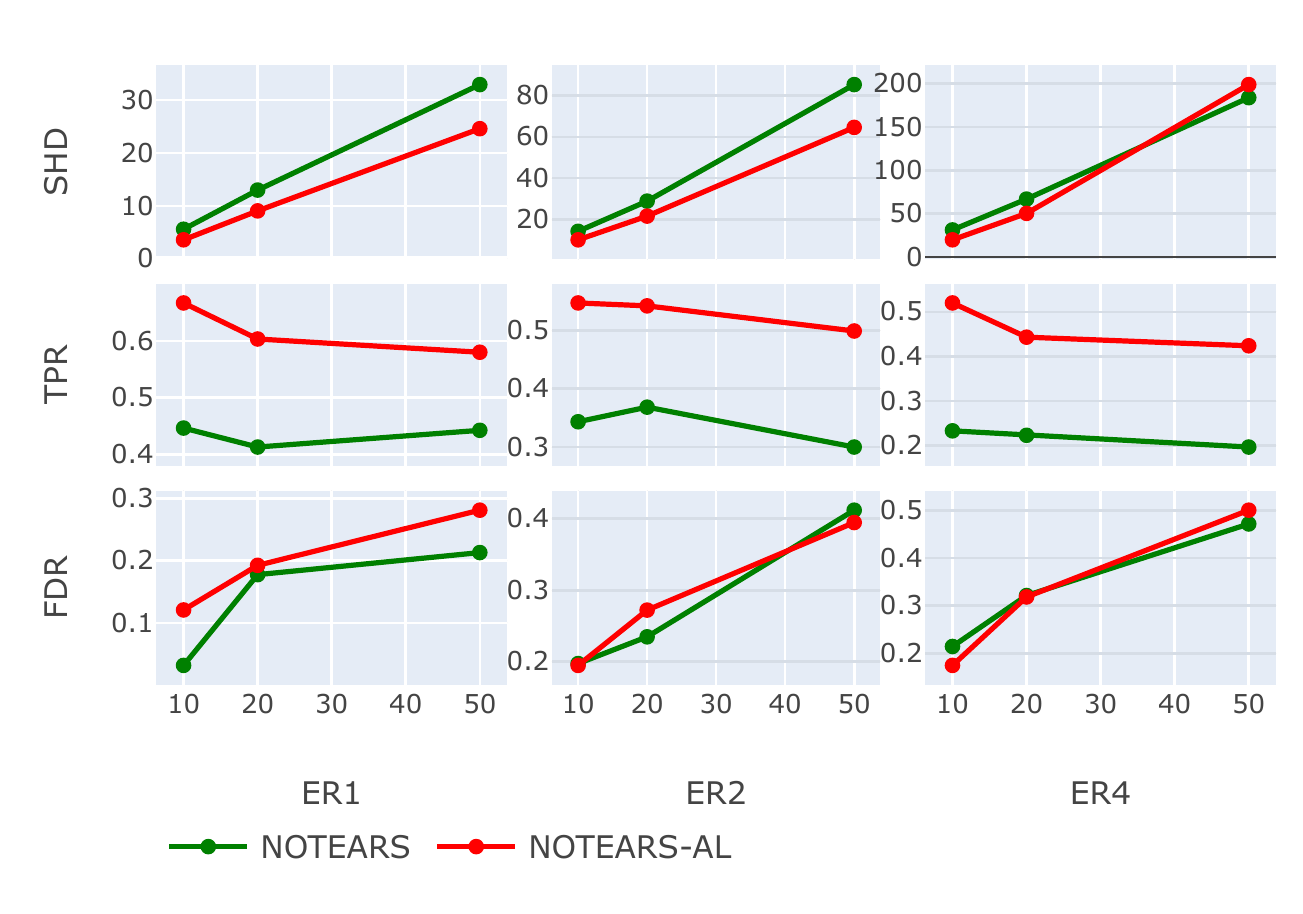} 
	\caption{Numerical results for logistic regression in terms of SHD, TPR, and FDR on ER graphs, with sample size $n = 1000$. For TPR, higher is better. For the other two metrics, lower is better. Columns: random graph types, ER$k$ means graphs with $k\times d$ expected edges. }
	\label{fig6}
\end{figure}

Figure \ref{fig6} shows the structure recovery results for $n=1000$. By comparison, We can find the TPR of NOTEARS-AL is significantly higher than that of NOTEARS with a fixed threshold. At the same time, in terms of the other two metrics, i.e., SHD and FDR, both methods keep at a similar level. In general, NOTEARS-AL consistently outperforms the original NOTEARS.  

\subsection{Real Data}
Finally, consistent with Zheng et al. \cite{zheng2018dags}, we compare our method to NOTEARS on the same real dataset, which records the expression levels of $11$ phosphoproteins and phospholipids in human cells. This dataset is commonly used in the literature on Bayesian network structure inference as it provides a network that is widely accepted by the biological community. Based on $d = 11$ cell types and $n = 911$ observational samples, the ground truth structure given in \cite{sachs2005causal} contains $17$ edges. In our experiments, NOTEARS with $0.1$ threshold estimates $22$ edges with an SHD of $20$, compared to $20$ estimates for NOTEARS-AL with an SHD of $19$. 
\section{Conclusion}
\label{sec:conclusion}
	
	\noindent Based on the methodology of NOTEARS, we present a method for score-based learning of DAG models that do not need a hard thresholding post-processing step. The critical technical design is to use an adaptive penalty level for each edge rather than a common one. Notably, we prove that our method enjoys the oracle properties when the adaptive penalties satisfy some mild conditions. With a suitable choice of hyper-parameter in the second step, where we apply a modified $K$-fold cross-validation to handle, our method can zero out edges automatically in each iteration. Our simulation experiments show the effectiveness of our method, and it is worth emphasizing that the adaptive strategy in our method is generally more flexible than the hard-threshold strategy since it is a purely data-driven procedure, which enables our method to own the ability to adapt to different types of graphs. 

\bibliography{ref}
\bibliographystyle{IEEEtran}

\section{Proof of Theorems}
\label{append_proof}
\subsection{Proof of Theorem 1}
\noindent Let $\eta_n=n^p$, $p<0$, $\boldsymbol{\delta}=\begin{pmatrix}
\delta_{1,1} & \delta_{1,2} & \cdots & \delta_{1,d} \\
\delta_{2,1} & \delta_{2,2} & \cdots & \delta_{2,d} \\
\vdots  & \vdots  & \ddots & \vdots  \\
\delta_{d,1} & \delta_{d,2} & \cdots & \delta_{d,d} 
\end{pmatrix}$, $||\boldsymbol{\delta}||\le d$, where $d$ is a constant.
\begin{equation}
	\nonumber
	\begin{aligned}
		D_n(\boldsymbol{\delta})=& L_n(W^*+\eta_n\boldsymbol{\delta})-L_n(W^*)\\
		=&l_n(W^*+\eta_n\boldsymbol{\delta})-l_n(W^*)+\\
		&\lambda_n \sum_i\sum_j c_{ij} \left\{ |W^*_{ij}+\eta_n\delta_{ij}|-|W^*_{ij}|\right\}+\\
		&\frac{\rho}{2}\left\{ |h(W^*+\eta_n\boldsymbol{\delta})|^2-|h(W^*)|^2\right\}+\\
		&\alpha\left\{ h(W^*+\eta_n\boldsymbol{\delta})-h(W^*)\right\}
	\end{aligned}
\end{equation}
where $l_n(W)=\frac{1}{2n}||X-XW||_2^2$.

Let $a_{n,ij}=\lambda_n c_{ij}$, $(i,j)\in \mathcal{A}$, we can derive
\begin{equation}
	\nonumber
		\begin{aligned}
			D_n(\boldsymbol{\delta})\ge
			&l_n(W^*+\eta_n\boldsymbol{\delta})-l_n(W^*)+\\
			&\sum_{(i,j)\in \mathcal{A}} a_{n,ij} \left\{ |W^*_{ij}+\eta_n\delta_{ij}|-|W^*_{ij}|\right\}+\\
			&\frac{\rho}{2}\left\{ |h(W^*_{ij}+\eta_n\delta_{ij})|^2-|h(W^*_{ij})|^2\right\}+\\
			&\alpha\left\{ h(W^*_{ij}+\eta_n\delta_{ij})-h(W^*_{ij})\right\}\\
			\ge & l_n(W^*+\eta_n\boldsymbol{\delta})-l_n(W^*)+\\
			&\eta_n \sum_{(i,j)\in \mathcal{A}} a_{n,ij} |\delta_{ij}|+\\
			&\frac{\rho}{2}\left\{ |h(W^*+\eta_n\boldsymbol{\delta})|^2-|h(W^*)|^2\right\}+\\
			&\alpha\left\{ h(W^*+\eta_n\boldsymbol{\delta})-h(W^*)\right\}\\
			=&\nabla l_n(W^*)^T(\eta_n\boldsymbol{\delta})+\\
			&\frac{1}{2}(\eta_n\boldsymbol{\delta})^T[\nabla^2l_n(W^*)](\eta_n\boldsymbol{\delta})(1+o_p(1))-\\
			&\eta_n \sum_{(i,j)\in \mathcal{A}} a_{n,ij} |\delta_{ij}|+\\
			&\frac{\rho}{2}\left\{ |h(W^*+\eta_n\boldsymbol{\delta})|^2-|h(W^*)|^2\right\}+\\
			&\alpha\left\{ h(W^*+\eta_n\boldsymbol{\delta})-h(W^*)\right\}\\
			\triangleq & \rom{1}+\rom{2}+\rom{3}+\rom{4}
		\end{aligned}
	\end{equation}
where we define 
	\begin{equation}
	\nonumber
		\begin{aligned}
			\rom{1}=&\nabla l_n(W^*)^T(\eta_n\boldsymbol{\delta})\\
			\rom{2}=&\frac{1}{2}(\eta_n\boldsymbol{\delta})^T[\nabla^2l_n(W^*)](\eta_n\boldsymbol{\delta})(1+o_p(1))\\
			\rom{3}=&\eta_n \sum_{(i,j)\in \mathcal{A}} a_{n,ij} |\delta_{ij}|\\
			\rom{4}=&\frac{\rho}{2}\left\{ |h(W^*+\eta_n\boldsymbol{\delta})|^2-|h(W^*)|^2\right\}+\\
			&\alpha\left\{ h(W^*+\eta_n\boldsymbol{\delta})-h(W^*)\right\}
		\end{aligned}
	\end{equation}
For $\rom{1}$,
\begin{equation}
	\nonumber
		\begin{aligned}
			\rom{1}=&\nabla l_n(W^*)^T(\eta_n\boldsymbol{\delta})\\
			=&\eta_n\left(\nabla l_n(W^*)^T\right)\boldsymbol{\delta}\\
			=&\eta_n\left(\frac{1}{n}(X-XW^*)^TX^T\right)\boldsymbol{\delta}\\
			=&-\frac{1}{\sqrt{n}}\eta_n(\sqrt{n}\frac{1}{n}(X-XW^*)^TX^T)\boldsymbol{\delta}\\
			=&-O_p(\frac{1}{\sqrt{n}}\eta_n)\boldsymbol{\delta}.
		\end{aligned}
	\end{equation}
For $\rom{2}$,
\[
\rom{2}=\frac{1}{2}\eta_n^2\left\{\boldsymbol{\delta}^T[\frac{1}{n}X^TX]\boldsymbol{\delta} \right\}(1+o_P(1))>0.
\]
For $\rom{3}$,
\begin{equation}
	\nonumber
		\begin{aligned}
			\rom{3}=&-\eta_n \sum_{(i,j)\in \mathcal{A}} a_{n,ij} |\delta_{ij}|\\
			\ge &-\eta_n A n^q d,
		\end{aligned}
\end{equation}
where $A=\max\{a_{n,ij}/n^q: (i,j)\in \mathcal{A}\}$, and $A\in O(1)$.

\noindent For $\rom{4}$, since $h(W)\ge0$, $\forall W\in \mathbb{R}^{d\times d}$, we can derive
\[
\rom{4}=\left[ h(W^*+\eta_n\boldsymbol{\delta})-h(W^*)\right]\left\{\frac{\rho}{2}\left[ h(W^*+\eta_n\boldsymbol{\delta})+h(W^*)\right]
\right\}
\]
$\Rightarrow \qquad \rom{4}\ge 0$.

Thus, 
\begin{equation}
	\nonumber
		\begin{aligned}
            D_n(\boldsymbol{\delta})\ge & -O_p(\frac{1}{\sqrt{n}}\eta_n)\boldsymbol{\delta}+\frac{1}{2}\eta_n^2\left\{\boldsymbol{\delta}^T[\frac{1}{n}X^TX]\boldsymbol{\delta} \right\}(1+o_p(1))\\
            &-\eta_n A_n n^q d +\rom{4}
\end{aligned}
\end{equation}

We can find when $p>-\frac{1}{2}$ and $q\le -\frac{1}{2}$, \rom{2} dominates \rom{1} and \rom{3}, then, $D_n(\boldsymbol{\delta})>0$.

Thus, we can conclude that there exists a local minimizer of $L_n(W)$ with optimal convergence rate $||\widehat{W_n}-W^*||=O_p(n^{-\frac{1}{2}})$\cite{choi2010variable}.

\subsection{Proof of Theorem 2}
\noindent Firstly, for \textbf{sparsity property}, it is sufficient to prove $\forall (i,j)\in \mathcal{A}_c$,
\begin{equation}
	\begin{cases}
	\frac{\partial L_n(W)}{\partial W_{ij}}|_{W=\widehat{W}}<0 \quad \text{ if } -\varepsilon_n<\widehat{W}_{ij}<0\\
	\frac{\partial L_n({W})}{\partial W_{ij}}|_{W=\widehat{W}}>0 \quad \text{ if } 0<\widehat{W}_{ij}<\varepsilon_n
	\end{cases}
\end{equation}
with probability tending to 1 where $\varepsilon_n=O(n^{-\frac{1}{2}})$.

Then, 
\begin{equation}
	\nonumber
		\begin{aligned}
            \frac{\partial L_n(\widehat{W})}{\partial W_{ij}}=& 
            -\frac{1}{n}X_{.i}^T(X_{.j}-XW_{.j}^*)+\sum_{k=1}^{d}\frac{1}{n}X_{.i}^TX_{kj}(\widehat{W}_{ij}-W^*_{kj})\\
            &+\lambda_n c_{ij} sign\{\widehat{W}_{ij}\}+2\widehat{W}_{ij}(\rho h(\widehat{W})+\alpha)(e^{\widehat{W}\circ \widehat{W}})^T_{ij}
\end{aligned}
\end{equation}
where we can find
\begin{equation}
	\nonumber
		\begin{aligned}
            &\frac{1}{n}X_{.i}^T(X_{.j}-XW_{.j}^*)=O_p(n^{-\frac{1}{2}}),\\
            &\sum_{k=1}^{d}\frac{1}{n}X_{.i}^TX_{kj}(\widehat{W}_{ij}-W^*_{kj})=O_p(n^{-\frac{1}{2}}),\\
            &2\widehat{W}_{ij}(\rho h(\widehat{W})+\alpha)(e^{\widehat{W}\circ \widehat{W}})^T_{ij}
\begin{cases}
            <0 \quad \text{ if } -\varepsilon_n<\widehat{W}_{ij}<0\\
            >0 \quad \text{ if } 0<\widehat{W}_{ij}<\varepsilon_n
\end{cases}
\end{aligned}
\end{equation}
Let $b_{n,ij}=\lambda_n c_{ij}$. Therefore, for $\forall (i,j)\in \mathcal{A}_c$, if $\sqrt{n} b_{n,ij}\rightarrow \infty$, then, the sign of $\frac{\partial L_n(\widehat{W})}{\partial W_{ij}}$ is dominated by $sgn\{\widehat{W}_{ij}\}$.
\\
\\
\noindent Secondly, for \textbf{asymptotic normality property}, we concentrate on set $\mathcal{A}_i$, $i=1,2,...,d$, and use $W_{\mathcal{A}_i}$ to denote the sub adjacency matrix of graphs, $X_{\mathcal{A}_i}$ to denote the sub observation data of corresponding variables, and $C_{\mathcal{A}_i}$ to denote the corresponding adaptive penalty weights.

Denote 
\begin{equation}
	\nonumber
		\begin{aligned}
            &\rom{1}(W)=\frac{1}{2n}||X-XW||^2_2\\
            &\rom{2}(W)=\lambda_n\sum_i\sum_j|c_{ij}W_{ij}|\\
            &\rom{3}(W)=\frac{\rho}{2}|h(W)|^2+\alpha h(W).
\end{aligned}
\end{equation}
Then, 
\[
\nabla_{\mathcal{A}_i}L_n(\widehat{W}_{\mathcal{A}_i})=\nabla_{\mathcal{A}_i}\rom{1}(\widehat{W}_{\mathcal{A}_i})+\nabla_{\mathcal{A}_i}\rom{2}(\widehat{W}_{\mathcal{A}_i})+\nabla_{\mathcal{A}_i}\rom{3}(\widehat{W}_{\mathcal{A}_i}).
\]
By Taylor expansion at $\widehat{W}_{\mathcal{A}_i}=W^*_{\mathcal{A}_i}$, we have
\begin{equation}
	\nonumber
		\begin{aligned}
        \nabla_{\mathcal{A}_i}\rom{1}(\widehat{W}_{\mathcal{A}_i})=&\nabla_{\mathcal{A}_i}\rom{1}(W^*_{\mathcal{A}_i})+\left[\nabla^2_{\mathcal{A}_i}\rom{1}(W^*_{\mathcal{A}_i})\right](\widehat{W}_{\mathcal{A}_i}-W^*_{\mathcal{A}_i})\\
        =&\frac{1}{n}X^T_{\mathcal{A}_i}X_{\mathcal{A}_i}(\widehat{W}_{\mathcal{A}_i}-W^*_{\mathcal{A}_i})-\frac{1}{n}X^T_{\mathcal{A}_i}(X_{\mathcal{A}_i}-X_{\mathcal{A}_i} W^*_{\mathcal{A}_i}),\\
        \nabla_{\mathcal{A}_i}\rom{2}(\widehat{W}_{\mathcal{A}_i})=&\lambda_n C_{\mathcal{A}_i} sgn(\widehat{W}_{\mathcal{A}_i})\\
        =&\lambda_n C_{\mathcal{A}_i} sgn(W^*_{\mathcal{A}})+(\widehat{W}_{\mathcal{A}_i}-W^*_{\mathcal{A}_i})o_p(1),\\
        \nabla_{\mathcal{A}_i}\rom{3}(\widehat{W}_{\mathcal{A}_i})=&(\rho |h(\widehat{W}_{\mathcal{A}_i})|+\alpha)\nabla h(\widehat{W}_{\mathcal{A}_i})\\
        =& (\widehat{W}_{\mathcal{A}_i}-W^*_{\mathcal{A}_i})o_p(1).
\end{aligned}
\end{equation}
Therefore, now we have
\begin{equation}
	\nonumber
		\begin{aligned}
        \nabla_{\mathcal{A}_i}L_n(\widehat{W}_{\mathcal{A}_i})=&\frac{1}{n}X^T_{\mathcal{A}_i}X_{\mathcal{A}_i}(\widehat{W}_{\mathcal{A}_i}-W^*_{\mathcal{A}_i})-\frac{1}{n}X^T_{\mathcal{A}_i}(X_{\mathcal{A}_i}-X_{\mathcal{A}_i} W^*_{\mathcal{A}_i})\\
        &+\lambda_n C_{\mathcal{A}_i} sgn(W^*_{\mathcal{A}_i})+(\widehat{W}_{\mathcal{A}_i}-W^*_{\mathcal{A}_i})o_p(1)\\
        &+(\widehat{W}_{\mathcal{A}_i}-W^*_{\mathcal{A}_i})o_p(1)\\
        =&0.
\end{aligned}
\end{equation}
From Theorem 1, we have $a_{n,ij}=o_p(n^{-\frac{1}{2}})$, $\forall (i,j) \in \mathcal{A}$, and $||\widehat{W}_{\mathcal{A}}-W^*_{\mathcal{A}}||=O_p(n^{-\frac{1}{2}})$, then, 
\[
        \frac{1}{n}X^T_{\mathcal{A}_i}X_{\mathcal{A}_i}(\widehat{W}_{\mathcal{A}_i}-W^*_{\mathcal{A}_i})-\frac{1}{n}X^T_{\mathcal{A}_i}(X_{\mathcal{A}_i}-X_{\mathcal{A}_i} W^*_{\mathcal{A}_i})+o_p(n^{-\frac{1}{2}})=0.
\]
\begin{equation}
	\nonumber
		\begin{aligned}
\Rightarrow \sqrt{n}(\widehat{W}_{\mathcal{A}_i}-W^*_{\mathcal{A}_i})=&\left(\frac{1}{n}X^T_{\mathcal{A}_i}X_{\mathcal{A}_i}\right)^{-1}\times\\
&\left[\frac{1}{\sqrt{n}}X^T_{\mathcal{A}_i}(X_{\mathcal{A}_i}-X_{\mathcal{A}_i}W^*_{\mathcal{A}_i})-o_p(1)\right]
\end{aligned}
\end{equation}
Therefore, under condition~2, by Lindeberg-Feller central limit theorem in linear regression case, as our design satisfies the following mild condition\cite{van2000asymptotic} 
\[
\max_{1\le k \le n}|(X_{\mathcal{A}_i})_k|=o(n^{1/2}),
\]
\[
\forall i\in{1,2,...,d}, \sqrt{n}(\widehat{W}_{\mathcal{A}_i}-{W^*}_{\mathcal{A}_i})\xrightarrow{d}X\sim N(0,\sigma^2 D_{i0}^{-1}).
\]

\end{document}